\documentclass[runningheads]{llncs}

\usepackage{graphicx}

\usepackage{algorithm}
\usepackage[noend]{algpseudocode}
\usepackage{caption}
\usepackage{subcaption}
\usepackage{amsmath}
\usepackage{siunitx}
\usepackage{graphicx}
\usepackage{amssymb}
\usepackage{comment}

\usepackage{helvet} 
\usepackage{courier}  

\usepackage{color, soul}

\begin{document}
\title{Transferring Domain Knowledge with an Adviser in Continuous Tasks}
%
%
\author{Rukshan Wijesinghe\inst{1,2}\and
Kasun Vithanage\inst{2}\and
Dumindu Tissera\inst{1,2}\and
Alex Xavier\inst{2} \and
Subha Fernando\inst{2}\and
Jayathu Samarawickrama\inst{1,2}} 

\authorrunning{R D Wijesinghe et al.}
%
\institute{Department of Electronic and Telecommunication Engineering, University of Moratuwa, Sri Lanka \and
CODEGEN QBITS Lab, , University of Moratuwa, Sri Lanka }
\maketitle              
\begin{abstract}

Recent advances in Reinforcement Learning (RL) have surpassed human-level performance in many simulated environments. However, existing reinforcement learning techniques are incapable of explicitly incorporating already known domain-specific knowledge into the learning process. Therefore, the agents have to explore and learn the domain knowledge independently through a trial and error approach, which consumes both time and resources to make valid responses. Hence, we adapt the Deep Deterministic Policy Gradient (DDPG) algorithm to incorporate an adviser, which allows integrating domain knowledge in the form of pre-learned policies or pre-defined relationships to enhance the agent's learning process. Our experiments on OpenAi Gym benchmark tasks show that integrating domain knowledge through advisers expedites the learning and improves the policy towards better optima.

\label{se:abs}

\vspace{-0.2in}
\end{abstract}

\section{Introduction}
\label{se:intro}
\vspace{-0.12in}
Conventional reinforcement learning approaches have been limited to domains with low dimensional discrete state and action spaces or fully observable state and action spaces, where handcrafted features are heavily used. But with the emergence of deep Q-networks (DQN) \cite{DQN}, its applicability was extended to high dimensional state spaces. DQN has been able to surpass human-level performance in some of the challenging Atari 2600 games using only unprocessed pixels as input \cite{DQN}. This was still not a generalized solution and DQN was not suited well for the higher dimensional or continuous action spaces \cite{continuous_control,dpg}. The Deep Deterministic Policy Gradient algorithm (DDPG) which is derived from Deterministic Policy Gradient \cite{dpg} extended the Deep Q learning for continuous state and action space.

Although these advancements of RL have reached continuous state and action spaces they are still unable to directly incorporate already known domain knowledge into the learning process. Thus, they unnecessarily consume time and computational resources to acquire the fundamental known knowledge through learning it from scratch, i.e., agents will follow a trial and error approach many times before successfully converging to an optimal policy. In a simulated world, this is not efficient and it is particularly not welcomed in the real world tasks where the agents cannot make fatal mistakes during learning such as the autonomous navigation domain. To this end, an algorithm which facilitates to incorporate domain knowledge enables an agent to accelerate the learning procedure by limiting the exploration space and converge to better policies. 

In this paper, we propose a novel approach to train an agent efficiently in continuous and high-dimensional state-action spaces. Our approach adapts the DDPG algorithm to incorporate already available information to the training process in the form of an adviser to accelerate the learning process. The DDPG algorithm updates the policy in each iteration with approximated policy gradients which are derived from the gradients of the critic network's output with respect to the parameters of the actor network. However, this approach updates the policy parameters directly and does not facilitate the use of domain knowledge for the policy updating process. In contrast, we update the existing policy to a new policy based on a two-step approach. During the policy parameter update in each iteration, we first set a temporary target to the policy and then push the current policy towards that target by reducing the L2 distance between the current and the target policies.

This two-fold optimization facilitates taking the adviser's suggestions into account when updating the policy. In addition, the adviser can be used to enforce the agent to explore better regions of the state-action space to extract better policies while reducing the exploration cost. We theoretically prove the convergence of the adapted DDPG algorithm and empirically show that the proposed approach itself improves over the existing DDPG algorithm in chosen benchmark tasks in the continuous domain. We further plug advisers to the adapted DDPG algorithm to show accelerated learning, validating the utility of the two-fold policy updating process.

\section{Related Work}
\label{se:related_work}
\vspace{-0.12in}
Modern foundations of reinforcement learning are formed as a result of intertwining several paths of trial and error methods and the solution to the problem of optimal control with temporal methods \cite{sutton1998introduction}. Initially, the applications of RL were restricted to the low dimensional discrete action-state spaces. Deep Q Networks \cite{DQN} extended its applicability to the continuous high dimensional state spaces. Later, the DDPG algorithm \cite{continuous_control,dpg} combined the DQN and deterministic policy gradient algorithm to handle continuous high dimensional state-action spaces simultaneously. Integrating DDPG with actor-critic \cite{Actor-Critic} architecture allows learning parameterized continuous policies, optimized with policy gradient which is derived in terms of parameterized Q-value function. 

Reinforcement learning has been an emerging trend in the autonomous navigation domain \cite{kang2019generalization,kahn2017uncertainty,mirowski2018learning,wayne2018unsupervised}. End-to-end trained asynchronous deep RL \cite{asynchronous} based models were used to do continuous control of mobile robots in map less navigation \cite{Autonomous_navigation}. Recently there is an increased concern on reducing training time, increasing sample efficiency, and minimizing the trial and error nature of the learning process to make RL applicable to actual navigational applications more safely and confidently. DQN and DDPG are highly sample-inefficient since they demand a large number of samples during the training. Nagabandi \textit{et. al} \cite{nagabandi} proposed combining medium-sized neural networks with model predictive control and a model-free learner initialized by deep neural network dynamic models that were tested on MuJoCo locomotion tasks \cite{mujoco} that achieved high sample efficiencies. Kahn \textit{et.al} \cite{computatio_graph} proposed a self supervising generalized computational graph for autonomous navigation which subsumes the advantages of both model-free and model-based methods. They have empirically shown that their model surpasses the performance of Double Q-network which is an enhanced version of DQN that gives less overestimates and variations\cite{double_Q}. 

Transferring learned knowledge between problem instances or separate agents reduces the trial and error nature of the training process. Successor-feature-based reinforcement learning \cite{successor} has been used to do such knowledge transfers across similar navigational environments. Taylor \textit{et.al} \cite{representationtransfer} introduced complexification and offline RL algorithms for transferring knowledge between agents with different internal representations. Multitask and transfer learning has been utilized in autonomous agents where they can learn multiple tasks at once and apply the generalized knowledge to new domains \cite{actor-mimic}. Ross \textit{et.al} \cite{ross2011reduction} discuss the DAgger algorithm which is similar to no-regret online learning algorithms. It uses a dataset of trajectories collected using an expert, to initialize policies that can mimic the expert better. Methods for automatically mapping between different tasks through analyzing agent experience have shown to increase the training speed of reinforcement learning \cite{autonomoustransferlearning}.

Self-imitation learning \cite{oh2018self} learns to reproduce the past good decisions of the agent to enhance deep exploration. Hindsight experience replay has been used with DDPG to overcome exploration bottlenecks on simulated robotics tasks in \cite{nair2018overcoming}. Hester \textit{et.al} \cite{hester2018deep} proposed deep Q learning from demonstrations that use small sets of demonstration data to accelerate the learning process. The effect of function approximation errors in actor-critic settings has been addressed by employing a novel variant of  Double Q-Learning \cite{fujimoto2018addressing}. Maximum entropy reinforcement learning is used in off-policy actor-critic methods \cite{actor_critic} to overcome the issues of sample inefficiency and convergence in conjunction \cite{haarnoja2018soft}. Continuous variants of Q-learning combined with learned models have shown to be effective in addressing the sample complexity of RL in continuous domains \cite{gu2016continuous}.

Our approach uses the actor-critic architecture deviates from the existing methods due to several reasons. First, we adapt the DDPG algorithm  to incorporate domain knowledge as an adviser in continuous tasks with high dimensional state-action spaces. Secondly, we employee the adviser in data collection to enforce the agent to explore regions of state-action space with higher return.

\section{Method}
\label{se:method}
\vspace{-0.12in}
 The proposed adapted DDPG algorithm improves the policy in the direction of the gradient of the Q-value function. It also facilitates integrating pre-learned policies or existing relationships as advisers to transfer domain knowledge. During the training process, advisers can be deployed in; 1) data collection, as well as 2) policy updating processes. Once the adviser is involved in the data collection process, it enforces the agent to explore better regions of state and action spaces according to the adviser's perspective. When the adviser is incorporated into the policy updating process, it aids to reach better policies rapidly by selecting the best set of actions. In this section, we first introduce the adapted DDPG algorithm and prove its convergence. Then we explicitly describe the proposed ways of employing an adviser for the data collection and policy updating processes to achieve an efficient training approach in continuous tasks. 
\vspace{-0.1in}

\subsection{Adapted Deep Deterministic Policy Gradient Algorithm}
\label{se:AVG}

\vspace{-0.2in}
\begin{figure}[H]
	\centering
	\includegraphics[width=0.35\linewidth]{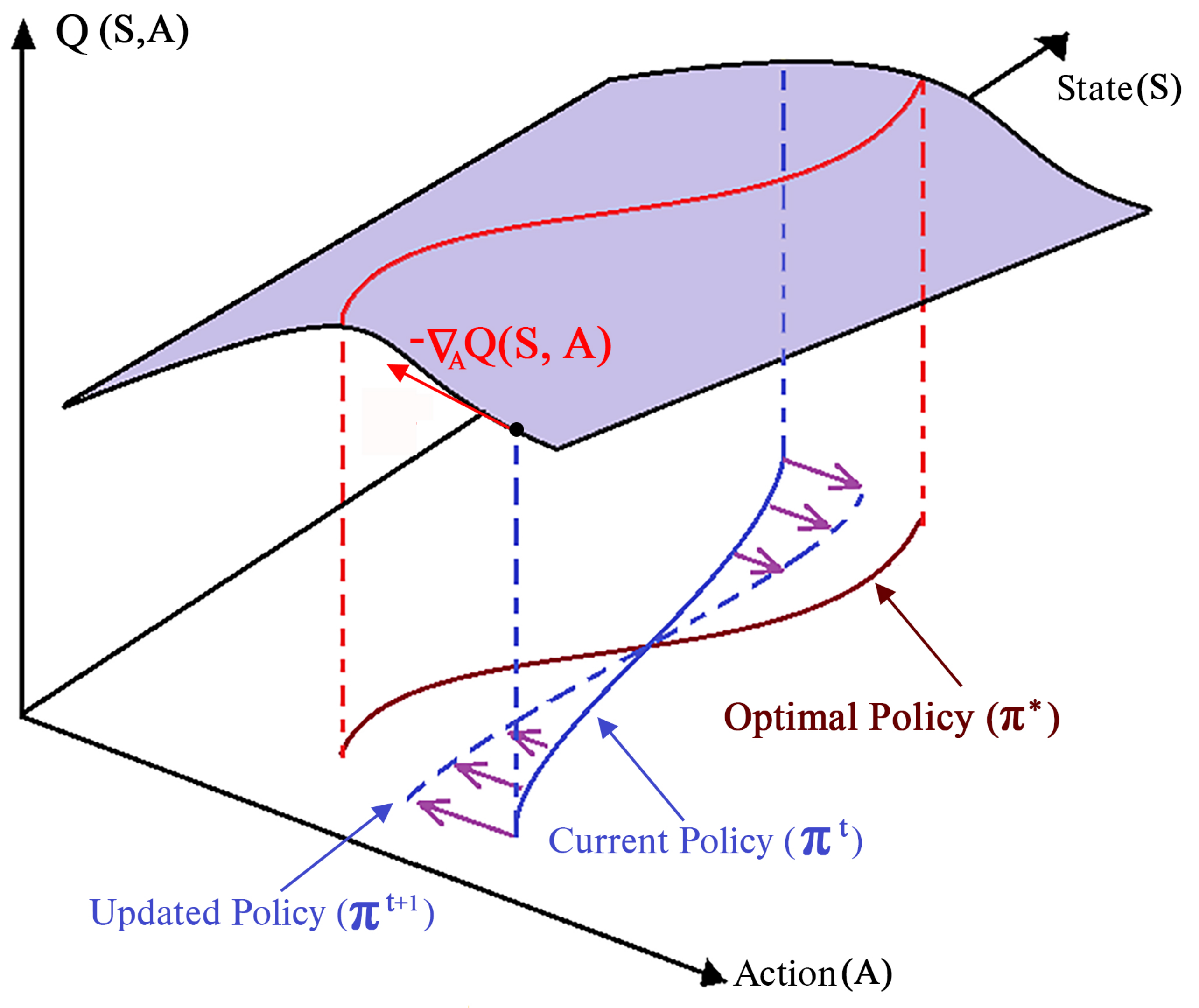}
	\caption{Policy updating method with gradient of action-value function.}
\label{Figure01}
\end{figure}
\vspace{-0.2in}

To extract the optimal policy in the continuous state-action space, moving the existing policy in the direction of the gradient of the Q-value function is computationally efficient than globally maximizing the Q-value function \cite{dpg}. Therefore, the proposed method can be utilized to extract better policies in tasks with high-dimensional continuous state and action spaces, by improving the current policy in the direction of the gradient of the action-value (Q-value) function. The surface in Figure \ref{Figure01} represents the Q-value function corresponding to a hypothetical RL problem in continuous domain. For explanation simplicity, it only contains a single action variable ($A$) and a single state variable ($S$). Let the $\pi(S;\phi)$ be the policy function which governs the actions in given states. $\pi$ is parameterized by $\phi$ which are known as policy function parameters.

The Peak red line of the surface represents the Q-values of the state-action pairs in optimal policy $\pi^*(S)$ corresponding to an instance. Its projection on the state-action plane denotes the optimal policy. The direction of the gradient of the Q-value function with respect to actions $\nabla_A Q(S,A)$ corresponding to a point on current policy $\pi^t(S)$ at a particular time step $t$, always leads towards either local or global optimal-policy. Therefore, the term $\nabla_A Q(S,A)$ can be used to update the current policy and obtain a better updated policy $\pi^{t+1}(S)$ by pushing $\pi^t$ in the direction of $\nabla_A Q(S,A_{\pi^t})$ as shown in Figure \ref{Figure01}. Thus, the corresponding policy improvement at a particular step can be represented by;
\vspace{-0.08in}
\begin{equation}
\label{equation1}
    \pi^{t+1}(S) \gets \pi^t(S) + \beta \nabla_A Q(S,A_{\pi^t}).
    \vspace{-0.08in}
\end{equation}
Here, $\beta$ is the updating rate of the current policy, which represents the degree of the shift between the updated and current policies.
Once the updated policy $\pi^{t+1}(S)$ is calculated, it is used as a temporary target to optimize the current policy $\pi^t(S;\phi)$. We update the policy function parameters $\phi$ by performing a gradient descent to minimize the loss $L_\pi$;
\vspace{-0.08in}
\begin{equation}
\label{equation2}
    L_\pi = \frac{1}{n}\sum[\pi^{t+1}(S)-\pi^t(S;\phi)]^2,
    \vspace{-0.08in}
\end{equation}
which is the mean squared error between current and updated policy samples. The main advantage of this two-fold policy update is that we can plug suggestions of an adviser who has domain knowledge, in between the aforementioned two steps to achieve a better-updated policy at a particular time step.
\vspace{-0.1in}
\begin{algorithm}[H]
	\caption{Adapted DDPG Algorithm}\label{algorithm1}

		For each update of actor network $\pi(S;\phi)$ and critic network $Q(S,A;\theta)$ at a given time step $t$ ;
	\begin{algorithmic}[1]
	
		 \State  Select batch of experiences $M = <S,A,R,S'>$ randomly from memory replay  buffer with the size of $n$ 
		\State Set $\hat{Q}(S,A) \gets R + \gamma Q^-(S',\pi^-(S';\phi^-))$
		\State  Update $\theta$ by minimizing the loss function
		
		$L_Q = \frac{1}{n}\Sigma(\hat{Q}(S,A)-Q(S,A;\theta))^2$
		
		\State Set $\pi^{t+1}(S) \gets \pi^t(S;\phi) + \beta \nabla_A Q(S,A_{\pi^t};\theta)$
		
		\State Update $\phi$ by minimizing the loss function 
		
		$L_\pi = \frac{1}{n}\Sigma(\pi^{t+1}(S)-\pi^t(S;\phi))^2$
		
		\State Update target network parameters $\theta^-$ and $\phi^-$
		
		$\theta^- \gets \tau\theta +(1-\tau)\theta^-$
		
		$\phi^- \gets \tau\phi +(1-\tau)\phi^-$

	\end{algorithmic}
\end{algorithm}
\vspace{-0.1in}

In our approach, the Q-value function is updated with the temporal difference error similarly to the Deep Deterministic Policy Gradient (DDPG) algorithm presented in \cite{continuous_control}. It maintains two parameterized Q-value functions known as Q-network $Q(s,a;\theta)$ and target Q-network $Q^-(s,a;\theta^-)$. Similarly, it maintains two policy functions named as policy network $\pi(s,a;\phi)$ and target policy network $\pi^-(s,a;\phi^-)$. Algorithm \ref{algorithm1} illustrates the steps followed in each update of the Q-network and policy network. A soft update mechanism weighted by $\tau$ and $(1-\tau)$ (where $0<\tau << 1$) is used to update the target policy network and target Q-network as shown in the last step of Algorithm\ref{algorithm1}. Maintaining two separate networks and soft updating mechanisms enhance the stability of the learning process and it supports training the critic network without divergence \cite{continuous_control}.

\subsection{Convergence of Adapted DDPG Algorithm}
\vspace{-0.1in}

\subsubsection{Smooth Concave Functions}

If $f : \mathbb{R}^n \xrightarrow{} \mathbb{R}$  is a twice differentiable function and holds Lipschitz continuity with constant $L > 0$ then,
\vspace{-0.08in}
\begin{equation}
\label{eqn1}
\left\|\nabla f(y) - \nabla f(x) \right\| \leq L \left\|x - y \right\|  \; \forall x, y \in \mathbb{R}^n 
\vspace{-0.08in}
\end{equation}
Here,  $L$ is a measurement for the smoothness of the function \cite{GD} and if $f(x)$ is a concave function then the following inequality holds;
\vspace{-0.08in}
\begin{equation}
\label{eqn2}
f(y) \geq f(x) + \langle\nabla f(x),(y-x)\rangle + \frac{L}{2} \left\|y - x \right\|^2 \; \forall x, y \in \mathbb{R}^n 
\vspace{-0.08in}
\end{equation}
Here, $\langle\:,\:\rangle$ denotes the standard inner product between two vectors \cite{GD}.

\subsubsection{Proof of the Convergence}
\label{proof}
\vspace{-0.1in}

Let  $Q : \mathbb{R}^{m+n} \xrightarrow{} \mathbb{R}$ be the action value function where $m$ and $n$ represents number of state variables and number of actions respectively. $S$ and $A$ be the state and action vectors respectively such that $S \in \mathbb{R}^m$ and $A \in \mathbb{R}^n$. Let's define $(S, A ) = (s_1, s_2, \ldots, s_m, a_1, a_2, \ldots, a_n)$ where $s_1,  \ldots, s_m $ and $a_1, \ldots, a_n \in \mathbb{R}$. Let $A_\pi = \pi(S)$ where $\pi$ is the policy function that predicts the action to be executed for a given state $S$. Once the policy update at a particular time step $t$ is considered;
\vspace{-0.08in}
\begin{equation}
\label{eqn3}
A_{\pi^{t+1}} = \pi^{t+1}(S) = \pi^{t}(S) + \beta \nabla_AQ(S,A_{\pi^t})
\vspace{-0.08in}
\end{equation}
If Q(S, A) is a concave and twice differentiable function with Lipschitz  continuity then considering the equation \ref{eqn2};
\vspace{-0.08in}
\begin{equation}
\label{eqn4}
\begin{aligned}
Q(S,A_{\pi^{t+1}}) \geq Q(S,A_{\pi^t}) + \langle\nabla Q(S,A_{\pi^t}),((S,A_{\pi^{t+1}})-(S,A_{\pi^t}))\rangle \\
 -\frac{L}{2} \left\|(S,A_{\pi^{t+1}}) - (S,A_{\pi^t}) \right\|^2  
\end{aligned}
\vspace{-0.08in}
\end{equation}
\vspace{-0.08in}
\begin{equation}
\label{eqn5}
\begin{aligned}
Q(S,A_{\pi^{t+1}}) \geq Q(S,A_{\pi^t}) + \langle\nabla_A Q(S,A_{\pi^t}),(A_{\pi^{t+1}}-A_{\pi^t})\rangle \\
- \frac{L}{2} \left\|(A_{\pi^{t+1}}-A_{\pi^t}) \right\|^2
\end{aligned}
\vspace{-0.08in}
\end{equation}
Since $A_{\pi^{t+1}}-A_{\pi^t} = \beta \nabla_AQ(S,A_{\pi^t})$  by substituting for Equation \ref{eqn5}
\vspace{-0.08in}
\begin{equation}
\label{eqn7}
Q(S,A_{\pi^{t+1}}) \geq Q(S,A_{\pi^t}) + \beta\left\| \nabla_AQ(S,A_{\pi^t}) \right\|^2 - \frac{L\beta^2}{2} \left\| \nabla_AQ(S,A_{\pi^t}) \right\|^2
\vspace{-0.08in}
\end{equation}
Considering Equation \ref{eqn7} and if $0 <\beta \leq \frac{2}{L}$ then,
\vspace{-0.08in}
\begin{equation}
\label{eqn8}
Q(S,A_{\pi^{t+1}}) - Q(S,A_{\pi^t})\geq   \beta(1-\frac{\beta L}{2})\left\| \nabla_AQ(S,A_{\pi^t}) \right\|^2 \geq 0  
\vspace{-0.08in}
\end{equation}
Therefore, $Q((S,A_{\pi^{t+1}}) \geq Q(S,A_{\pi^t})$ at any time step. Once  all the time steps from $0$ to $k$ considered in Equation \ref{eqn8},
\vspace{-0.08in}
\begin{equation}
\label{eqn9}
\sum_{t=0}^{t=k} \{Q(S,A_{\pi^{t+1}}) - Q(S,A_{\pi^t})\}\geq \sum_{t=0}^{t=k}  \{\beta(1-\frac{\beta L}{2})\left\| \nabla_AQ(S,A_{\pi^t}) \right\|^2\}
\vspace{-0.08in}
\end{equation}
\vspace{-0.08in}
\begin{equation}
\label{eqn10}
Q(S,A_{\pi^{k+1}}) - Q(S,A_{\pi^0})\geq \beta(1-\frac{\beta L}{2}) \sum_{t=0}^{t=k}  \left\| \nabla_AQ(S,A_{\pi^t}) \right\|^2
\vspace{-0.08in}
\end{equation}
If $Q(S,A_{\pi^*})$ is the Q-value at optimal policy $\pi^*(S)$ then $Q(S,A_{\pi^*})\geq Q((S,A_{\pi^{k+1}})$. By considering Equation \ref{eqn10},
\vspace{-0.08in}
\begin{equation}
\label{eqn12}
\begin{aligned}
Q(S,A_{\pi^*}) - Q(S,A_{\pi^0})\geq Q((S,A_{\pi^{k+1}}) - Q(S,A_{\pi^0})\geq  \\ \beta(1-\frac{\beta L}{2}) \sum_{t=0}^{t=k}  \left\| \nabla_AQ(S,A_{\pi^t}) \right\|^2
\end{aligned}
\vspace{-0.08in}
\end{equation}
Equation \ref{eqn12} implies that as $k \xrightarrow{}{\infty}$, the right hand side of the inequality converges. 
\vspace{-0.08in}
\begin{equation}
\label{eqn14}
\therefore \lim_{k\to\infty} \left\| \nabla_AQ(S,A_{\pi^k}) \right\|^2 = 0
\vspace{-0.08in}
\end{equation}
Since $ \pi^*(S) = \arg\max_{A} Q(S,A)$, at optimal policy $\nabla_A Q(S,A_{\pi^*}) = 0$. Therefore, the equation \ref{eqn14} implies that $\lim_{k\to\infty} A_{\pi^{k+1}} = A_{\pi^*}$. 

\subsection{Adapted DDPG with Actor-Critic Agent}
\vspace{-0.2in}
\begin{figure}[H]
	\centering
	\includegraphics[width=0.4\linewidth]{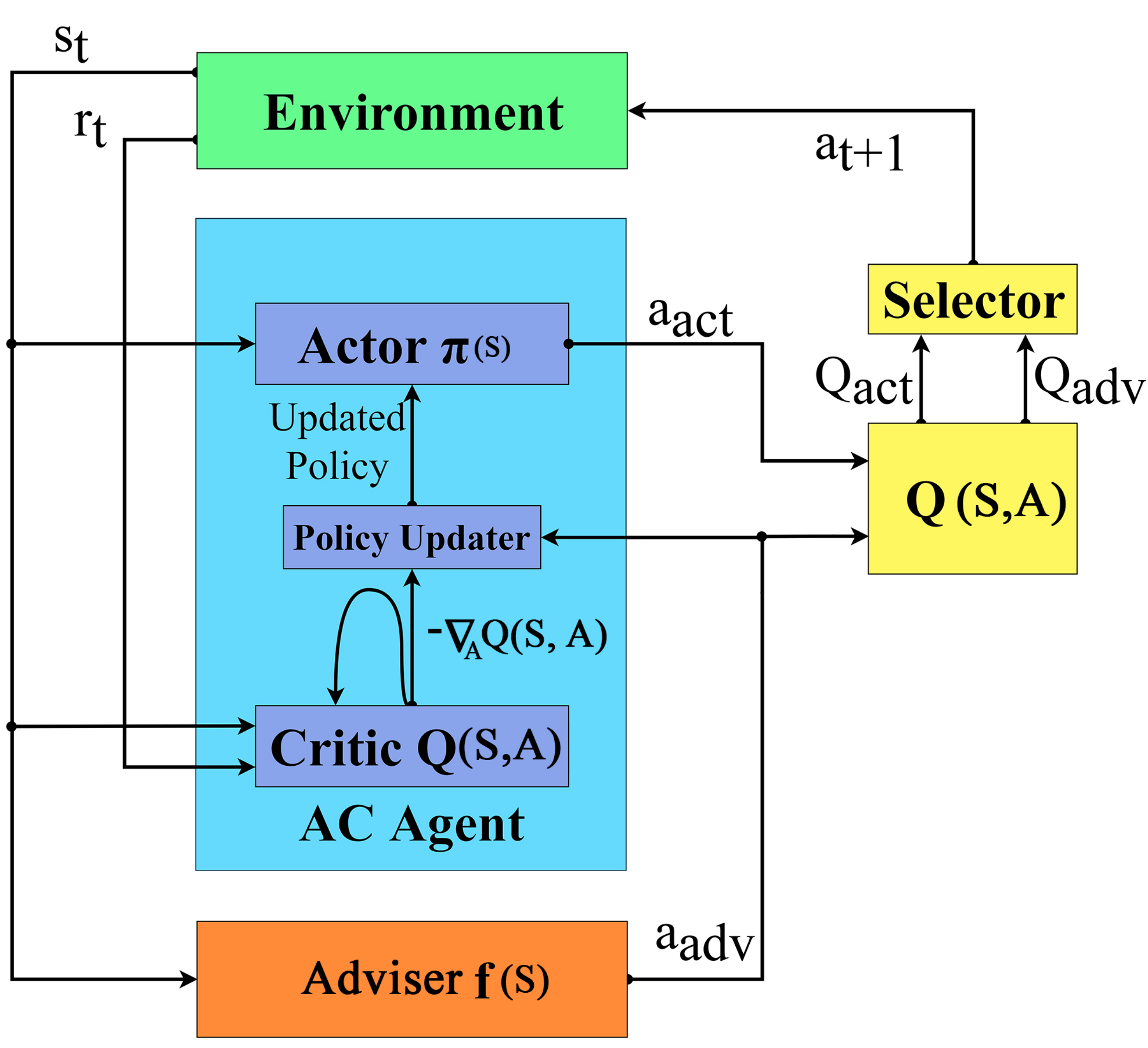}
	\caption{Actor-Critic Architecture with Adviser module.}
\label{Figure02}
\end{figure} 

\vspace{-0.2in}

Actor-Critic architecture is a well-known architecture used in model-free reinforcement learning. It consists of two main components, namely the Actor and the Critic. The actor decides which action to be taken by the agent and critic always tries to improve the performance of the actor by analyzing rewards received in each time-step. Generally, the actor is updated with the policy gradient approach, while the critic gets updated with the temporal difference error \cite{Actor-Critic}. Our agents are based on Actor-Critic architecture and learn a diverse set of benchmark tasks in the continuous domain. Figure \ref{Figure02} illustrates the basic block diagram of the Actor-Critic architecture that employs the adviser for both data collection and policy updating processes. In our implementation, the actor and critic modules represent the parameterized policy and Q-value functions respectively.

\subsection{Employing Adviser to Transfer Domain Knowledge } 
\vspace{-0.1in}

Although model-free value-based approaches have demonstrated state-of-the-art performance in the RL domain, the low sample efficiency is one of the main concerns which limits their applicability in real-world applications \cite{computatio_graph}. One solution to this issue is integrating domain knowledge into the learning process. Thus, the agent does not need to learn everything from scratch, and it directly affects the training efficiency and performance of task execution. Here, through an adviser, we propose two such techniques to integrate domain knowledge to support the learning process. Firstly, we enforce the agent to explore better regions in the state and action spaces by enabling adviser suggestions for the data collecting process. Secondly, we enhance the policy updating process by allowing the adviser to adjust the updated policy of each step to a better policy according to the current Q-Value knowledge.

\vspace{-0.2in}
\begin{minipage}{0.45\textwidth}
\begin{algorithm}[H]
	\caption{Data collection with adviser}\label{algorithm2}

		At a given time step $t$;
		
	\begin{algorithmic}[1]
		\State Observe current state $s_t$
		\State $a_{adv} \gets f(s_t)$
		\State $a_{act} \gets \pi(s_t)$
		\State $C \gets 1-e^{-\lambda N}$
		\State $\epsilon\gets \frac{e^{-\frac{Q(s_t, a_{avd})}{T}}}{e^{-\frac{Q(s_t, a_{avd})}{T}} + e^{-C\frac{Q(s_t, a_{act})}{T}}}$

		\State With probability $\epsilon$, $a_{t+1} \gets a_{adv}$
		\State Otherwise, $a_{t+1} \gets a_{act}$
		\State $a_{t+1} \gets a_{t+1} + noise$
        \vspace{0.035in}
	\end{algorithmic}
\end{algorithm}
\end{minipage}
\hfill
\begin{minipage}{0.45\textwidth}
\begin{algorithm}[H]
	\caption{Policy updating with adviser}\label{algorithm3}

	For each update of actor network $\pi(s;\phi)$ and critic network $Q(s,a;\theta)$ ;

\begin{algorithmic}[1]

	\State Steps 1-3 of algorithm\ref{algorithm1}
	\State $A_{adv} \gets f(S)$
	\State $A_{act} \gets \pi(S;\phi)$
	
	\State $\hat{A}(S) \gets A_{act} + \beta \nabla_A Q(S,A;\theta)$
	
	\For{$1:i:n$}
	\If {$Q(s^i,a^i_{adv}) > Q(s^i,\hat{a}^i)$} 
	\State $\hat{a}^i \gets a^i_{adv}$
	\EndIf
	\EndFor

	\State Steps 5-6 of algorithm\ref{algorithm1}

	\end{algorithmic}
\end{algorithm}
\end{minipage}

\subsection{Adviser for Data Collection Process}
\vspace{-0.1in}

Here, we employ an adviser $(f)$ which is a mapping of states ($S$) to actions ($A$), to make sampling more efficient by comparing actor's current predictions of actions with the adviser's suggestions. In each time step, both the actor ($A_{act}$) and the adviser ($A_{adv}$) suggests the action to be executed for the current state ($S_t$). Then, both suggestions are evaluated with respect to the current knowledge (Q-value function) of the critic module and the adviser's action is selected for the execution with a probability $\epsilon$. Calculation of $\epsilon$ is adapted by the work  \cite{fernandez2006probabilistic}. Our method deviates from theirs in several ways. We employ a version of softmax function to induce a higher probability corresponding to the action with higher Q-value. By varying the softmax temperature $T$, it is possible to change the priority given to the adviser. As shown in the Algorithm \ref{algorithm2} the constant $C$ ($C = 1- e^{-\lambda N}$) is a confidence value calculated on agent's behalf where $N$ is the number of episodes elapsed and $\lambda$ $(\lambda>0)$ is the decaying constant. This enforces the agent to give higher priority for adviser's suggestions at the beginning, enabling the agent to explore near a better policy, compared to the architecture without an adviser. In the end, we add a noise signal to the selected action for exploration. The noise generation is influenced by the Ornstein-Uhlenbeck process \cite{uhlenbeck}, and it is correlated with the input signal so that it ensures a better exploration near the selected action. 

\subsection{Adviser for Policy Updating Process}
\vspace{-0.1in}

Since the adapted DDPG algorithm improves the existing policy with an updated set of sampled actions, it enables integrating adviser's suggestions during the policy updating process as described in Algorithm \ref{algorithm3}. At each iteration, a batch of experiences with $n$ samples are fetched randomly from memory replay buffer, and Q-value function is updated as similar to Algorithm \ref{algorithm1}. Before updating the policy, both adviser suggestions and actor suggestions for the selected batch ($A_{adv}$ and $A_{act}$ respectively) are calculated. Thereafter, an updated set of actions ($\hat{A}(S)$) are calculated by considering action-value gradients corresponding to the actor's suggestions, similar to the Algorithm \ref{algorithm1}. In the next step, each updated action ($\hat{a}^i$) is replaced by the adviser's action ($a^i_{adv}$), if the Q-value corresponding to the adviser suggested action is greater than corresponding updated action ($\hat{a}^i$). In Algorithm \ref{algorithm3}, $s^i$ and $a^i$ refers to the state and action of the $i^{th}$ sample of the selected batch. Finally, policy parameters are updated with the modified set of updated actions corresponding to the selected batch.

\section{Experiments}
\label{se:exp}
\vspace{-0.12in}

To evaluate the performance of the adapted DDPG algorithm and adviser based agent architecture, we experiment on a diverse set of benchmark tasks in the continuous domain. This includes four OpenAI Gym \cite{OpenAI_Gym} environments namely,  Pendulum-v0, MountainCarContinuous-v0, LunarLanderContinuous-v2 and Bi- -pedalWalker-v2. For each task environment, we train three distinct agents using the DDPG, Adapted DDPG, and the Adapted DDPG with adviser algorithms in separation. We develop adviser modules using classical control approaches (Proportional Integral and Derivative controllers) and predefined rules for Pendulum, MountainCarContinuous, and  LunarLanderContinuous environments. For the BipedalWalker task, we deploy a pre-trained policy as the adviser. We employed neural networks to parameterize the the Q-value function and the policy function. Due to the lack of information about the smoothness of the Q-Value function, the updating rate $\beta$ of the adapted DDPG algorithm is set to a lower value  (0.01) to satisfy to satisfy the condition $\beta < \frac{2}{L}$ (see the Section \ref{proof}) to ensure the convergence.

\begin{table}[ht]
\vspace{-0.1in}
\centering
\caption{The averaged total episodes score of the trained agents for 30 runs with 500 eposides in each. The adapted DDPG surpasses the conventional DDPG in all tasks. Although the adviser performance is comparatively low, the adapted DDPG algorithm with an adviser shows the best performance.}
\label{Table01}
\begin{tabular}{l|c|c|c|c|}
\cline{2-5}
\multicolumn{1}{c|}{} &
  Pendulum &
  \begin{tabular}[c]{@{}c@{}}MountainCar\\ Continuous\end{tabular} &
  \begin{tabular}[c]{@{}c@{}}LunarLander\\ Continuous\end{tabular} &
  \begin{tabular}[c]{@{}c@{}}Bipedal Walker\end{tabular} \\ \hline
\multicolumn{1}{|l|}{adviser}                                                 & -508.5 & 12.2 & -126.5  & 20.1  \\ \hline
\multicolumn{1}{|l|}{DDPG}                                                    & -398.9 & 28.5 & -65.7 & 100.3 \\ \hline
\multicolumn{1}{|l|}{\begin{tabular}[c]{@{}l@{}}Adapted \\ DDPG\end{tabular}} & -272.7  & 55.1 & -31.4 & 150.1 \\ \hline
\multicolumn{1}{|l|}{\begin{tabular}[c]{@{}l@{}}Adapted \\ DDPG + adviser\end{tabular}} &
  \textbf{-178.3} &
  \textbf{93.4} &
  \textbf{-30.1} &
  \textbf{190.3} \\ \hline
\end{tabular}
\vspace{-0.1in}
\end{table}

In each task, we train the agents for 30 runs where each run consists of a pre-defined number of training episodes. After each run, we test the agents for 500 episodes. We define the total episode score as the total reward earned by the agent during all the steps in a given episode. We take the average of such total episode scores gained in all 500 test episodes in a given run and average this figure again over the 30 runs. Table \ref{Table01} reports this average total episode score where the adapted DDPG algorithm comfortably surpasses the conventional DDPG algorithm in all tasks. The adviser alone (see first raw of Table \ref{Table01}) is not capable of performing the given tasks to the level of DDPG or adapted DDPG trained agents. However, the combination of the adapted DDPG algorithm and the adviser attain the best performance in all the tasks. This shows that even though the adviser under-performs when deployed alone, it certainly assists the adapted DDPG agents to converge towards a policy with higher scores.

\vspace{-0.1in}

\begin{figure}[ht]
    \vspace{-0.1in}
	\centering
	\begin{subfigure}[b]{0.45\linewidth}
		\centering
		\includegraphics[width=0.9\linewidth]{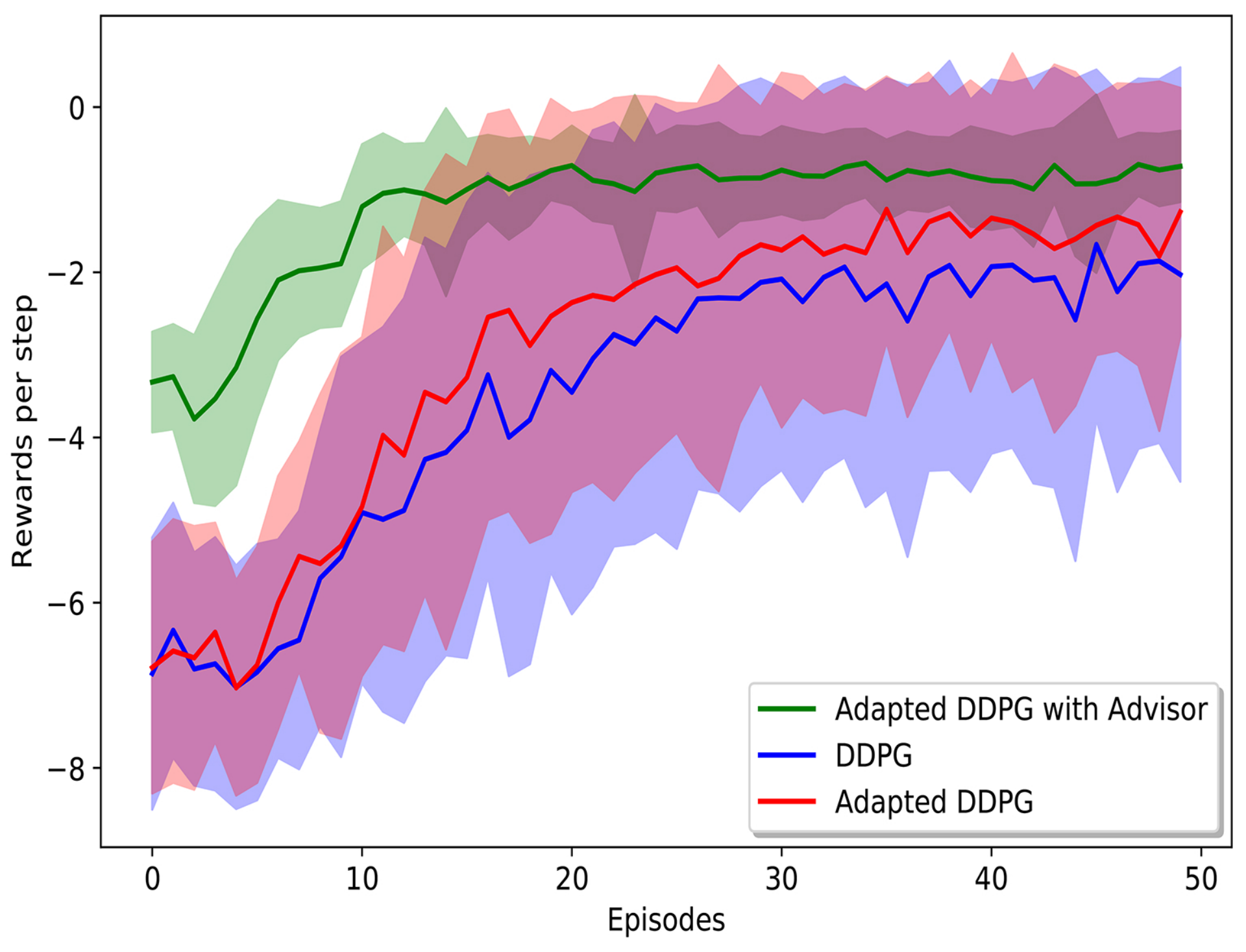}
		\caption{Pendulum}
		\label{Figure9_a}
	\end{subfigure}
	\begin{subfigure}[b]{0.45\linewidth}
		\centering
		\includegraphics[width=0.9\linewidth]{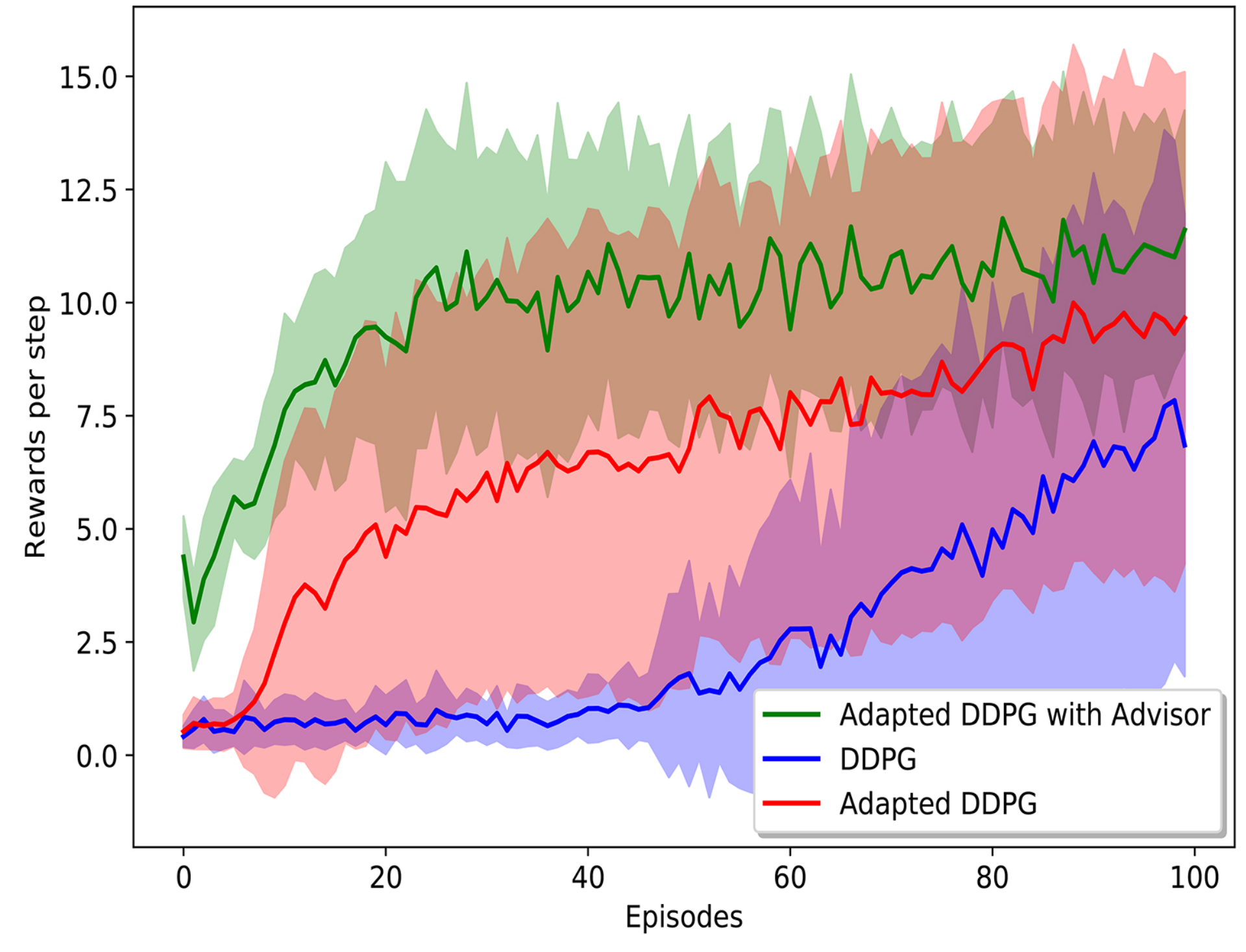}
		\caption{MountainCar}
		\label{Figure9_b}
	\end{subfigure}
	\newline
	\begin{subfigure}[b]{0.45\linewidth}
		\centering
		\includegraphics[width=0.9\linewidth]{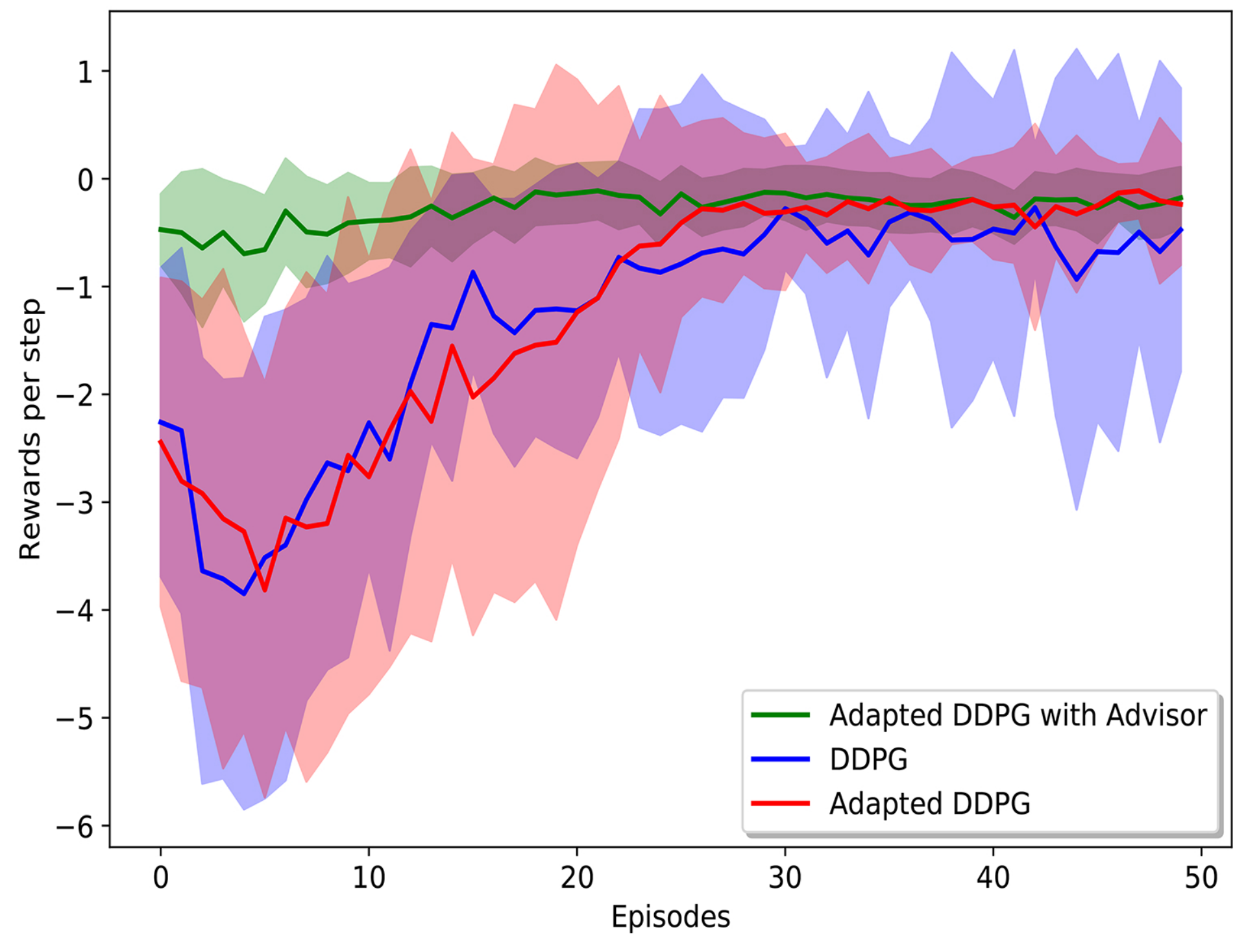}
		\caption{LunarLanderContinuous}
		\label{Figure9_c}
	\end{subfigure}
	\begin{subfigure}[b]{0.45\linewidth}
		\centering
		\includegraphics[width=0.9\linewidth]{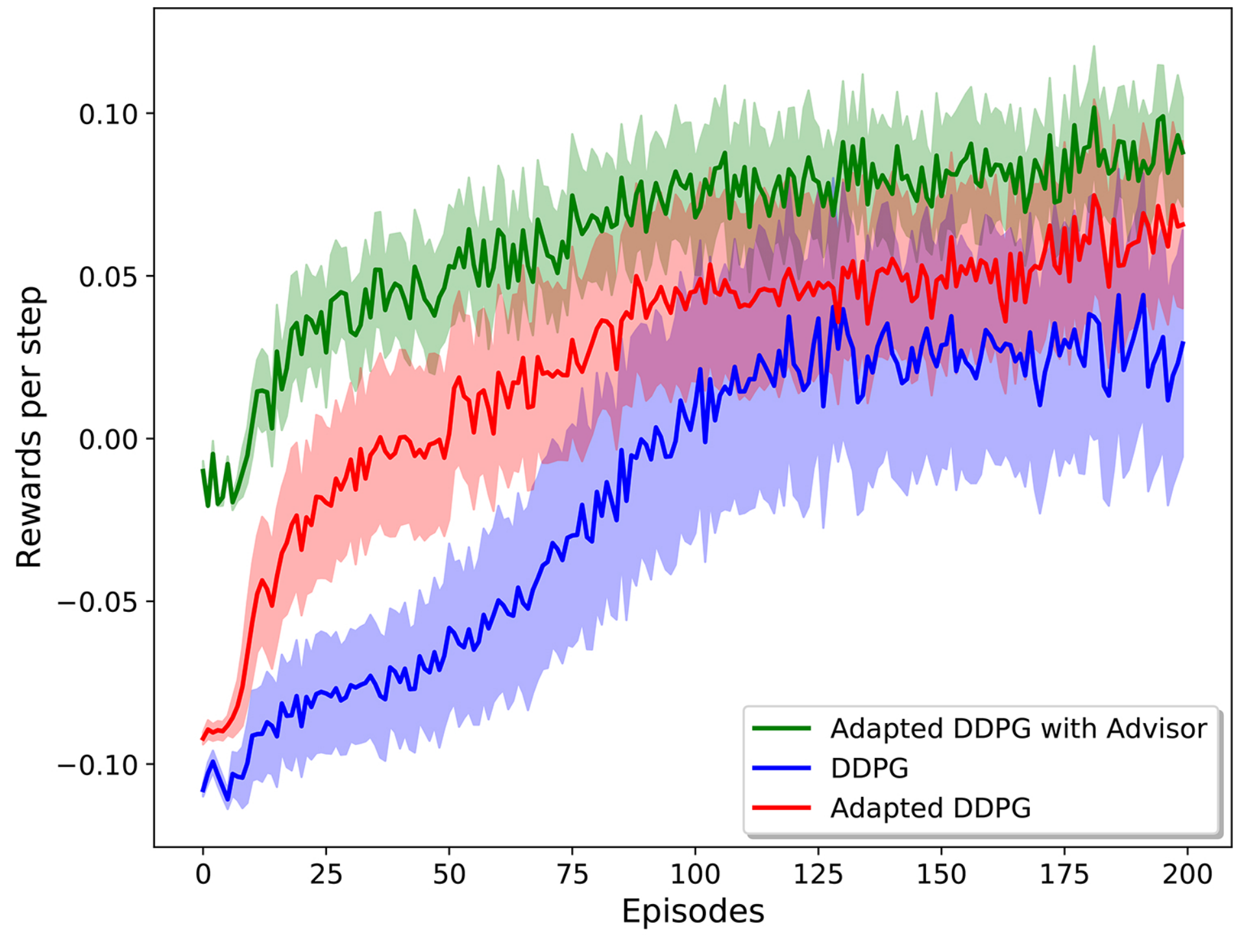}
		\caption{BipedalWalker}
		\label{Figure9_d}
	\end{subfigure}
	\caption{Reward per step of trained agents with the episode number on continuous benchmark tasks. The adapted DDPG algorithm reach higher reward levels rapidly compared to the DDPG algorithm, and adviser always accelerates the training speed further with a low variance in the learning curve.} 
	\label{Figure9}
	\vspace{-0.1in}
\end{figure} 

\vspace{-0.1in}

We further plot the averaged training reward per step, with the episode number in Figure \ref{Figure9}. For a given episode, the reward per step is the total reward earned by the agent during the episode divided by the number of steps.  We obtain this value for all episodes across all runs. The average reward per step for a particular indexed episode is calculated by averaging these values belonging to the same indexed episodes across the 30 runs. It demonstrates that the adapted DDPG algorithm achieves higher reward levels rapidly than conventional DDPG. It further illustrates that incorporating an adviser during the training phase expedites the learning process significantly compared to both DDPG and adapted DDPG algorithms. It is also evident that the agent with adviser converges to better policies, achieving higher rewards compared to other methods. Additionally, the adapted DDPG with adviser shows a considerably less variance than all the others.

\section{Conclusion}
\label{se:conclusion}
\vspace{-0.12in}

In this paper, we presented a novel approach of adapting the DDPG algorithm to incorporate an adviser that represents the domain knowledge to expedite the training process. The adviser in our actor-critic architecture causes the data collection and policy updating process to be more effective. We theoretically proved the convergence of the adapted DDPG algorithm and showed experimentally that the proposed adapted DDPG algorithm outperforms the standard DDPG algorithm in conventional RL benchmark tasks in the continuous domain. Additionally, we also demonstrated that the proposed two-fold policy updating mechanism of the adapted DDPG algorithm effectively incorporates core domain knowledge,  which is available as common relationships or set of rules to the training process, resulting in accelerated convergence towards better policies.

 \bibliographystyle{splncs04}
 \bibliography{references}
\end{document}